\documentclass[runningheads]{llncs}
\usepackage[T1]{fontenc}
\usepackage{amsmath}
\usepackage{xcolor}
\usepackage[
  colorlinks=true,
  citecolor=blue,
  linkcolor=blue,
  urlcolor=blue
]{hyperref}
\usepackage{booktabs}
\usepackage{comment}
\usepackage{glossaries}
\usepackage{amssymb}
\usepackage{amsmath}

\makeglossaries
\newglossaryentry{medsam}{name={MedSAM},description={Medical Segment Anything Model}}

\usepackage{graphicx,verbatim}
\begin{document}
\title{LoGSAM: Parameter-Efficient Cross-Modal Grounding for MRI Segmentation}
\titlerunning{LoGSAM; Automatic MRI Tumor Extraction}
%

\author{First Author\inst{1}\orcidID{0000-1111-2222-3333} \and
Second Author\inst{2,3}\orcidID{1111-2222-3333-4444} \and
Third Author\inst{3}\orcidID{2222--3333-4444-5555}}
\authorrunning{F. Author et al.}
%
\institute{Princeton University, Princeton NJ 08544, USA \and
Springer Heidelberg, Tiergartenstr. 17, 69121 Heidelberg, Germany
\email{lncs@springer.com}\\
\url{http://www.springer.com/gp/computer-science/lncs} \and
ABC Institute, Rupert-Karls-University Heidelberg, Heidelberg, Germany\\
\email{\{abc,lncs\}@uni-heidelberg.de}}

\author{Mohammad Robaitul Islam Bhuiyan\inst{1}\and Sheethal Bhat\inst{1,2}\and Melika Qahqaie\inst{1,2} \and Tri-Thien Nguyen\inst{1} \and Paula Andrea Pérez-Toro\inst{1} \and Tomás Arias-Vergara\inst{1}\and Andreas Maier\inst{1}}  
\authorrunning{M.R.I. Bhuiyan et. al} 
\institute{Pattern Recognition Lab, Friedrich-Alexander-Universität Erlangen-Nürnberg, Erlangen, Germany\and Siemens Healthineers, Erlangen, Germany \\
    \email{mohammad.ri.bhuiyan@fau.de}}

\maketitle              
\begin{abstract}
Precise localization and delineation of brain tumors using magnetic resonance imaging (MRI) are essential for planning therapy and guiding surgical decisions. To address this, we propose LoGSAM, a parameter-efficient, detection-driven framework that transforms radiologist dictation into text prompts for foundation-model-based localization and segmentation. Radiologist speech is first transcribed and translated using a pretrained Whisper ASR model, followed by negation-aware clinical NLP to extract tumor-specific textual prompts. These prompts guide text-conditioned tumor localization via a LoRA-adapted vision-language detection model, Grounding DINO (GDINO). The predicted bounding boxes are used as prompts for MedSAM to generate pixel-level tumor masks without any additional fine-tuning. On BRISC 2025, LoGSAM attains a Dice score of 80.32\%, reaching 98.6\% of a fully fine-tuned GDINO + MedSAM baseline while training fewer than 5\% of its parameters, indicating a favorable accuracy/parameter trade-off.  In addition, we evaluate the full pipeline using German dictations from a board-certified radiologist on unseen MRI scans, achieving 91.7\% case-level class-extraction accuracy. These results highlight the feasibility of constructing a modular speech-to-segmentation pipeline from pretrained foundation models with minimal parameter updates. 




\end{abstract}
\section{Introduction}
Brain tumors such as gliomas, meningiomas, and pituitary tumors represent serious pathologies that can impact the central nervous system and require precise characterization for treatment planning \cite{louis2016who}. Towards this end, contrast-enhanced T1-weighted magnetic resonance imaging (MRI) is a widely used imaging modality for identifying active tumor regions \cite{bauer2013survey}. Although MRI provides extensive diagnostic information, manual tumor delineation is time-consuming and depends on radiologist expertise \cite{chang2007radiotherapy}.  Despite recent advances, state-of-the-art (SOTA) segmentation algorithms require large datasets with dense pixel-level annotations. These annotations are labor-intensive and expensive to collect in practice \cite{bauer2013survey}. Moreover, in routine clinical workflows, radiologists typically dictate reports, which can be used as additional supervisory signals. However, due to ambiguity, variability, and negation in clinical language, naive keyword matching might lead to increased false positives \cite{jorg2023structured,chapman2001negex}. To address these gaps, we propose a parameter-efficient speech-to-segmentation framework that transforms radiologist dictations into structured prompts for tumor localization and segmentation. 

In the speech domain, high-accuracy ASR systems enable reliable transcription of multilingual radiologist dictations, making speech-derived supervision increasingly practical in clinical workflows \cite{radford2023robust}. 
Similarly, in the vision-language domain, clinical reports and textual cues have been successfully leveraged as weak supervisory signals for representation learning \cite{zhang2022contrastive,wang2022medclip}. Furthermore, grounding-based approaches connect textual phrases to image regions, enabling text-conditioned localization \cite{liu2024grounding}.
In parallel, prompt-based segmentation frameworks such as the Segment Anything Model (SAM) and its medical adaptation MedSAM have demonstrated strong generalization across modalities when provided with appropriate prompts \cite{kirillov2023segany,MedSAM}. However, adapting large vision–language and segmentation models to specific medical domains remains computationally expensive. Parameter-efficient fine-tuning (PEFT) methods, such as LoRA, enable targeted adaptation while keeping most parameters frozen \cite{hu2022lora}. Recent works combining LoRA-based grounding with SAM-style segmentation suggest that detection-driven prompting can effectively guide segmentation in specialized domains \cite{hu2024loramedsam,rasaee2025gdinoussam}. 

Building upon these advances, we propose LoGSAM, a parameter-efficient, detection-driven speech-to-segmentation framework for brain tumor analysis. Radiologist dictations are transcribed \cite{radford2023robust} and converted into structured tumor prompts using negation-aware NLP \cite{chapman2001negex}, which guides text-conditioned localization via a LoRA-adapted grounding model. The resulting bounding boxes condition a prompt-based segmentation model to produce pixel-level masks without additional fine-tuning.

\subsubsection{Main Contributions:} 

1) A unified speech-to-segmentation framework integrating ASR, clinical NLP, vision-language grounding, and medical segmentation.
2) Systematic analysis of LoRA adaptation for MRI-specific grounding, with ablations on rank and injection location.
3) Demonstration of robust out-of-distribution evaluation on 554 Kaggle MRI cases \cite{kaggle_mri_bbox} and radiologist-guided testing with German dictations, highlighting the practical applicability of the proposed speech-to-segmentation pipeline.

\section{Method}
\subsection{Datasets}
This study uses two datasets. BRISC 2025 includes 6,000 contrast-enhanced T1-weighted 2D MRI scans across four categories—\textit{Glioma}, \textit{Meningioma}, \textit{Pituitary}, and \textit{Healthy}—covering multiple anatomical planes (axial, coronal, and sagittal). All the images are in JPEG format with 512$\times$512 resolution, carefully curated, labeled, and quality‑checked by radiologists and physicians. The images are collected from multiple public datasets that lack comprehensive acquisition metadata such as patient identifiers, slice indices, scanner type, or sequence parameters ~\cite{fateh2026brisc}.The dataset is mildly imbalanced, with pituitary (29.28\%) and meningioma (27.25\%), followed by glioma (23.35\%) and healthy (20.12\%). The dataset is split into 4000 (train), 1000 (val), and 1000 (test) images.

As an out-of-distribution (OOD) dataset, we use the Kaggle \textit{MRI for Brain Tumor with Bounding Box}~\cite{kaggle_mri_bbox}, consisting of 554 JPEG images. The annotations were created with the \textit{LabelImg} tool in YOLO format, with most images labeled at a resolution of $512 \times 512$. No information on the patients' numbers, slices, or scanner type is available.


\subsection{LoGSAM Pipeline}

\begin{figure}
    \centering
    \includegraphics[width=\textwidth]{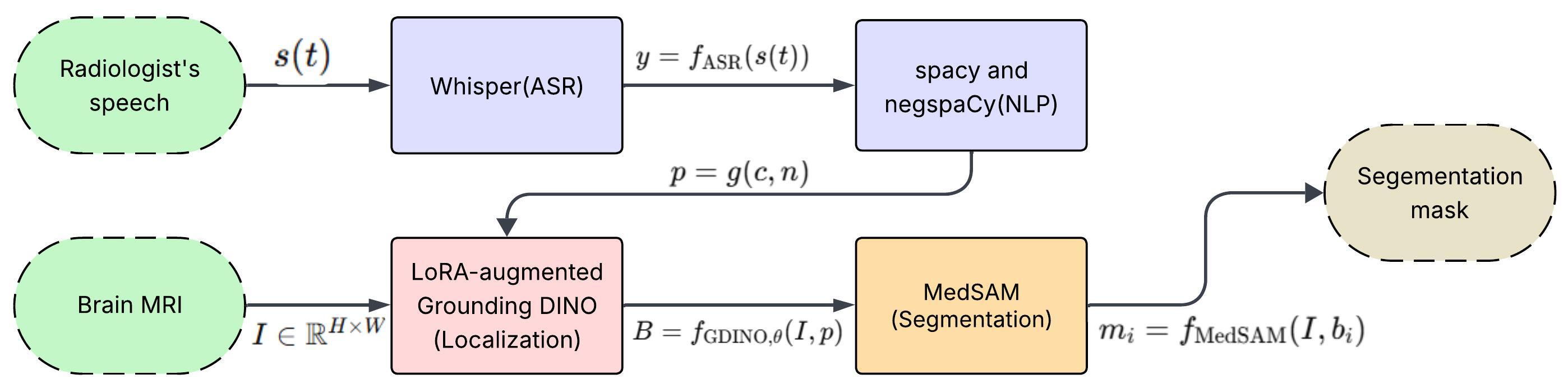} 
    \caption{\textbf{Overview of the proposed speech-guided detection-to-segmentation pipeline for brain tumor analysis.} The speech signal $s(t)$ is transcribed and translated using Whisper \cite{radford2023robust} ASR to obtain an English transcript $y = f_{\mathrm{ASR}}(s(t))$. The transcript is processed by spaCy \cite{honnibal2020spacy} + negspaCy \cite{chapman2001negex} to extract a tumor class cue and negation information, which are converted into a text prompt $p = g(c,n)$. In parallel, the T1-weighted brain MRI $I \in \mathbb{R}^{H \times W}$ is provided to a LoRA-augmented GDINO localizer, which combines $I$ and $p$ to predict bounding boxes $B = f_{\mathrm{GDINO},\theta}(I,p)$. Each predicted box $b_i \in B$ is then used as a prompt for MedSAM to generate a pixel-wise mask, yielding the final tumor segmentation output.}
    \label{fig:framework}
\end{figure}
As illustrated in Fig. \ref{fig:framework}, the proposed pipeline has three main components: (i) a speech-to-prompt generation module, (ii) a LoRA-augmented GDINO model for text-conditioned tumor localization, and (iii) a prompt-based segmentation model based on MedSAM.

\subsubsection{Speech-to-prompt generation:}
The input to this module is a board-certified radiologist’s dictated report, provided as an audio signal $s(t)$. The speech-to-prompt pipeline transforms this signal into a structured tumor prompt through the following processing chain:
\[
s(t)\xrightarrow{\,f_{\mathrm{ASR}}\,}y\xrightarrow{\,f_{\mathrm{NLP}}\,}(c,n)\xrightarrow{\,g\,}p,
\]
where, radiologist audio s(t) is transcribed via Whisper-large-v3 (5.7\% WER on German \cite{radford2023robust}), then processed by spaCy + negspaCy to extract tumor class $c \in \{\text{glioma}, \text{meningioma}, \text{pituitary}, \text{healthy}\}$ and negation indicator n. A synonym dictionary (e.g., glioblastoma → glioma) ensures clinical term normalization. Two-stage negation analysis detects global cues ('no evidence of tumor') and entity-level negations using NegEx-style rules\cite{chapman2001negex}, reducing false positives. Final prompt p encodes only class-level information.

\subsubsection{LoRA-augmented GDINO:}
\label{sssec:lora_config}
We adopt and expand the official GDINO implementation \cite{liu2024grounding} with a Swin-T backbone pretrained on ImageNet. The detection module $f_{\mathrm{GDINO},\theta}(I, p)$ serves as a text-conditioned tumor localizer, where $I$ is the input MRI slice and $p$ is the structured tumor prompt; GDINO encodes p with its BERT-base-uncased text encoder of 110 million parameters. The tumor classes are manually added into the text encoder to generate domain-specific prompt embeddings that GDINO can align with tumor regions. Additionally, we inject LoRA modules into selected attention and feed-forward layers for parameter-efficient MRI adaptation. Thus, low-rank update matrices are trained via parameters $\theta$, while most model parameters remain frozen. Specifically, LoRA modules are inserted into: 
(i) deformable self-attention layers in the encoder and decoder, 
(ii) vision–language cross-attention layers, 
(iii) feed-forward networks within transformer blocks, 
(iv) the initial layers of the bounding-box prediction head, 
(v) the feature enhancement module, and 
(vi) the final layers of the BERT text encoder.
This targeted adaptation strategy adapts the model to  MRI-specific tumor localization while maintaining parameter efficiency.

\subsubsection{MedSAM Segmentation:}
GDINO's predicted bounding boxes are used as prompts for MedSAM. 
Given an input image $I$, the LoRA-augmented localizer produces a set of candidate boxes
\begin{equation}
B=f_{\mathrm{GDINO},\theta}(I,p)=\{(b_i,s_i)\}_{i=1}^{N},
\label{eq:gdino_boxes}
\end{equation}
where $b_i$ denotes the $i$-th bounding box and $s\_i$ its confidence score. 
To suppress low-confidence predictions, we apply score-thresholding and retain
\begin{equation}
B_{\tau}=\{\,b_i \mid s_i \ge \tau\,\},
\label{eq:box_threshold}
\end{equation}
where $\tau$ denotes the confidence threshold. 
MedSAM is then used in inference mode with frozen weights and leverages its pretrained medical knowledge to produce a pixel-wise segmentation for each retained box:
\begin{equation}
m_i = f_{\mathrm{MedSAM}}(I,b_i), \qquad b_i\in B_{\tau},
\label{eq:medsam_mask}
\end{equation}
where $m_i\in\{0,1\}^{H\times W}$ denotes the predicted tumor/background mask. 
The final output of LoGSAM consists of the set of masks $\{m_i\}_{b_i \in B_{\tau}}$.
%

\subsubsection{Implementation Details: }

Radiologist dictations are transcribed using pretrained Whisper (large-v3) \cite{radford2023robust}. Tumor entities and negations are extracted using spaCy \cite{honnibal2020spacy} and negspaCy \cite{chapman2001negex} with a curated vocabulary. For localization, we use the official MMDetection \cite{chen2019mmdetection} for full fine-tuning and  expand the official implementation of GDINO \cite{liu2024grounding} for LoRA augmentation with a Swin-T backbone and default hyperparameters. Results are averaged over four runs under the same training set-up.  All experiments are implemented in PyTorch and run on single NVIDIA A100 GPU, equipped with 40GB of RAM. LoRA modules (rank $r=64$) are injected as described in Sec. \ref{sssec:lora_config}, and the model is trained on BRISC 2025 bounding-box annotations for 125 epochs using AdamW ($2\times10^{-4}$) with a OneCycle schedule of 10\% warmup and gradient clipping with a maximum norm of 5.0. For inference, given an MRI slice $I$ and text prompt $p$, this pretrained LoRA-augmented GDINO (\textit{Swin-T}) localizer predicts bounding boxes, and detections are filtered with a confidence threshold $\tau=0.3$. The threshold value was selected based on validation-set precision–recall trade-off analysis; lower values increased false-positive detections while higher values reduced recall on small tumors.  MedSAM (ViT-B) \cite{MedSAM}, pretrained on medical datasets, is used in frozen inference mode on these predicted bounding boxes to produce the segmentation masks.

\section{Results}


We evaluate our LoGSAM pipeline performance in two stages: (i) qualitative evaluation of the class extraction model and (ii) detection-to-segmentation performance.

\subsubsection{Speech-to-prompt generation module:}
Qualitative evaluation (Table~\ref{tab_prompt}) demonstrates correct prompt generation in 5 of 6 representative sample cases; one failure (case gg (227)) resulted from a transcription error (“glioblastoma” misrecognized as “glioplaston”). Performance is robust when tumor mentions are explicit and covered by the curated synonym list. The combined global negation detection and negspaCy gating mechanism reduces false-positive prompts. Overall, the module achieves reliable class extraction (91.7\% case-level accuracy) from real German clinical dictations, validating the feasibility of replacing manual prompt engineering with speech-driven prompting. 

\begin{table}[ht]
\centering
\caption{\textbf{Sample outputs of the Whisper and spaCy + negspaCy prompt-extraction stage}}\label{tab_prompt}
\begin{tabular}{lll}
\toprule
\textbf{Case\_id} &  \textbf{Class} & \textbf{Evidence} \\
\midrule
brisc2025\_test\_00013\_gl\_ax\_t1 & glioma  & glioblastoma \\



brisc2025\_test\_00430\_me\_co\_t1 & meningioma  & meningioma \\

brisc2025\_test\_00647\_no\_co\_t1 & healthy  & no target tumor term found \\

brisc2025\_test\_00686\_no\_sa\_t1 & healthy  & no evidence of diffusion \\

brisc2025\_test\_00743\_pi\_ax\_t1 & pituitary  & macroadenoma  \\

gg (227) (failure case) & healthy  & glioplaston \\ 

\bottomrule


\end{tabular}

\end{table}


\subsubsection{Detection-to-Segmentation Pipeline:}

The Detection-to-segmentation performance is reported as the mean over four runs under the same training configuration. Table~\ref{tab_dino} shows that  on the BRISC in-distribution dataset, LoRA-augmented GDINO retains 97.44\% of the fully fine-tuned model’s detection performance while updating only 4.96\% of parameters. The zero-shot baseline confirms that LoRA adaptation is essential; without it, detection degrades severely. Under OOD evaluation, LoRA-GDINO achieves a mAP@50 of 0.7955 and a mean IoU of 0.8239, indicating robust generalization with a moderate decrease relative to the fully fine-tuned baseline.

\begin{table}[ht]
\centering
\caption{\textbf{Detection performance of GDINO with and without LoRA augmentation under in-distribution and out-of-distribution (OOD) settings.}} \label{tab_dino}
\begin{tabular}{llll}
\toprule
Dataset &  Model & mAP@50 & Mean IoU \\
& & (Mean ± SD) & (Mean ± SD)\\
\midrule
BRISC in-distribution & Zero-shot GDINO & 0.4832 ± .040 & 0.5475 ± .030\\
 & Fully fine-tuned GDINO  & 0.8499 ± 0.020 & 0.9021 ± 0.015 \\
 & LoRA-augmented GDINO & 0.8282 ± 0.035 &  0.8631 ± 0.020 \\
\midrule
 Bbox OOD & Zero-shot GDINO & 0.5102 ± .035 & 0.6177 ± .020\\
& Fully fine-tuned GDINO & 0.8428 ± 0.030 & 0.8409 ± 0.020 \\
 & LoRA-augmented GDINO & 0.7955 ± 0.040 & 0.8239 ± 0.030 \\
\bottomrule
\end{tabular}
\end{table}

\begin{table}[ht]
\centering
\caption{\textbf{Segmentation performance on BRISC and Bbox datasets of the proposed LoGSAM and fully fine-tuned detection pipeline.}} \label{tab_seg}
\begin{tabular}{lll}
\toprule
Dataset &  Model & Mean Dice Score \\
& & (Mean ± SD)\\
\midrule
 BRISC in-distribution & Fully fine-tuned GDINO + MedSAM & 0.8145 ± .015   \\
 & Proposed LoGSAM & 0.8032 ± .020 \\
 \midrule
 Bbox OOD & Fully fine-tuned GDINO + MedSAM & 0.7923 ± .025   \\
 & Proposed LoGSAM & 0.7701 ± .030 \\
\bottomrule
\end{tabular}
\end{table}

Table~\ref{tab_seg} shows that the proposed LoGSAM framework achieved a mean Dice score of 0.8032 on BRISC, corresponding to 98.61\% of the fully fine-tuned GDINO + MedSAM (0.8145).  The advantage of LoGSAM lies in its substantially reduced parameter footprint  rather than in exceeding the baseline. Table~\ref{tab_seg} also confirms that the Dice drop from in-distribution to OOD is modest and consistent with the detection-level observations. Taken together, these results support the practical viability of parameter-efficient, detection-driven segmentation as a competitive alternative to full fine-tuning, while substantially reducing the number of trainable parameters.

Table~\ref{tab_class} represents that across all three tumor categories, LoGSAM achieves performance within 1.5 percentage points of the fully fine-tuned baseline, confirming that parameter-efficient adaptation does not disproportionately harm any single class. Healthy cases are excluded from Dice computation since no segmentation is produced for them.

\begin{table}[ht]
\centering
\caption{\textbf{Per-class Dice scores for the proposed LoGSAM and the fully fine-tuned baseline on the BRISC in-distribution test set.}}\label{tab_class}
\begin{tabular}{lllll}
\toprule
Model &  Glioma & Meningioma & Pituitary & Mean \\
\midrule
Fully fine-tuned GDINO + MedSAM & 0.798	& 0.825	& 0.821 & 0.8145  \\
Proposed LoGSAM  & 0.784 & 0.816 & 0.809 & 0.8032 \\

\bottomrule
\end{tabular}
\end{table}

\begin{figure}[ht]
    \centering
    \includegraphics[width=0.487\columnwidth]{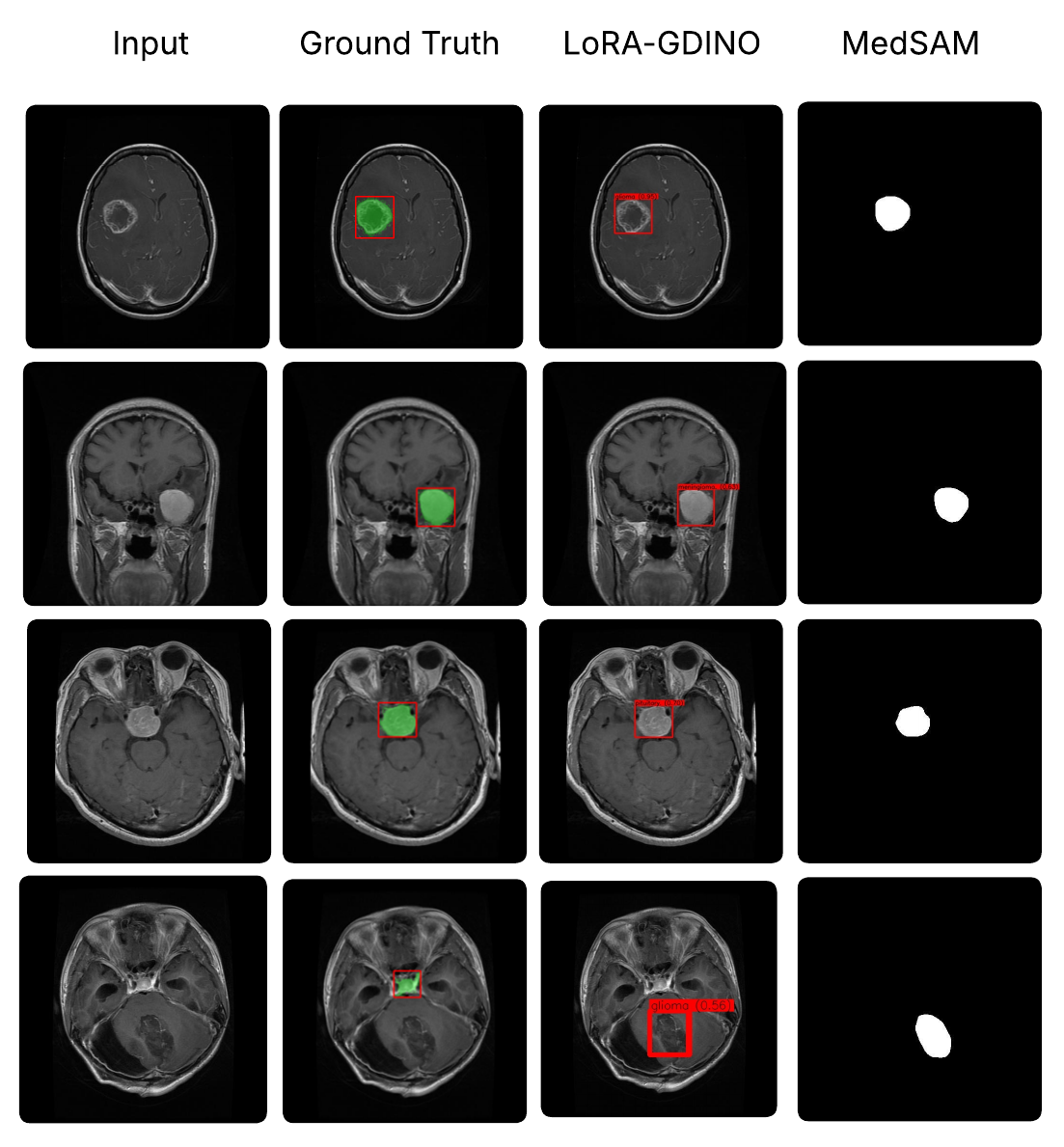}\hfill
    \includegraphics[width=0.493\columnwidth]{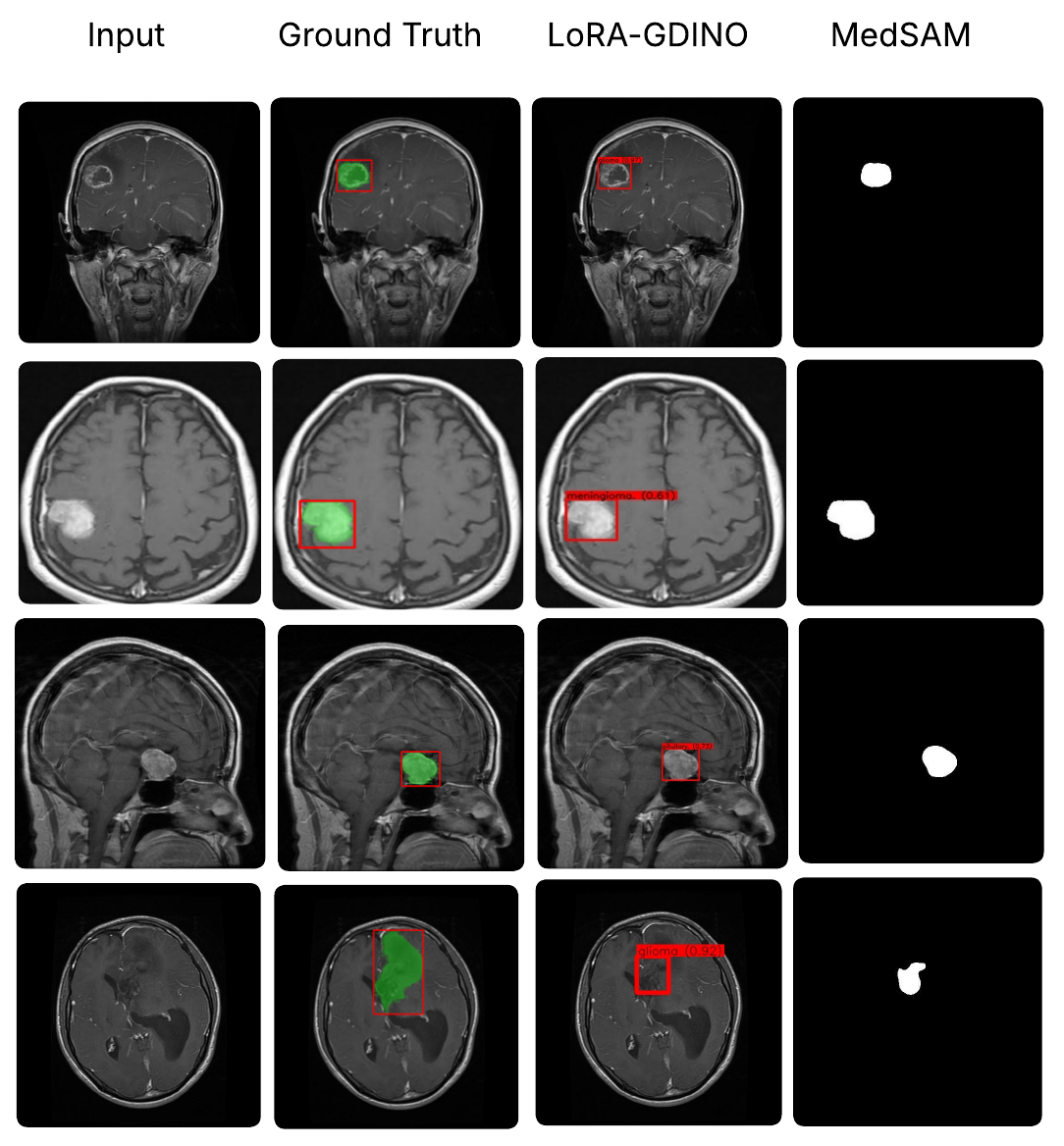}
     \caption{\textbf{Qualitative results at different stages of the proposed detection–to-segmentation pipeline.} The four rows show two cases each: glioma (row 1), meningioma (row 2), pituitary (row 3), and failure cases (row 4).}
    \label{fig:qualitative}
\end{figure}

\subsubsection{Qualitative Assessment:}
Fig.~\ref{fig:qualitative} shows in most of the cases the localizer identifies tumor regions accurately; the row-4 examples show two characteristic failures. The LoRA-GDINO completely missed the tumor in the left-side case and partially missed the tumor in the right-side case. The LoRA-augmented GDINO localizer accurately identifies tumor regions using text prompts, including challenging cases with low contrast or irregular boundaries. The resulting MedSAM masks closely align with ground-truth annotations, particularly in well-defined enhancing tumor cores. It also shows that the localization errors in LoRA-GDINO carry over to MedSAM masks. Overall, the masks are visually consistent with ground truth across the sample cases.

\subsubsection{Comparison:}
On BRISC 2025, LoGSAM attains a 0.8032 Dice score. It matches the multi-network ensemble baseline and approaches the fully supervised transformer SOTA on segmentation overlap. Yet it trains fewer than 5\% of parameters and adds a natural-language clinical interface. This demonstrates competitive, parameter-efficient tumor segmentation without additional fine-tuning of MedSAM.

\begin{table}[ht]
\centering
\caption{\textbf{Comparison of segmentation performance with related studies on the BRISC 2025 dataset.}}\label{tab_compare}
\begin{tabular}{lllll}
\toprule
Study &  Architecture & Metric  \\
\midrule
Fateh et al. (2026)\cite{fateh2026brisc} & SaberNet (Transformer) & Weighted mIoU: 80.6\%\\
Bamboo et al. (2025)\cite{bhamboo2025trustrefined} & Ensemble (UNet2.5D +  & Dice:0.798\\
&AttUNet2.5D + TransUNet)&\\
Proposed LoGSAM Table.~\ref{tab_seg} & GDINO and MedSAM (VLM) & Dice: 0.8032\\
\bottomrule

\end{tabular}
\end{table}

\begin{table}[ht]
\centering
\caption{\textbf{LoRA ablation on GDINO fine-tuning.} Reports the effect of LoRA rank size (32, 64, 128) and LoRA injection location at rank=64 on the trainable parameters and detection performance.}\label{tab1}
\begin{tabular}{llll}
\toprule
\textbf{LoRA Rank} & \textbf{Params (\%)} & \textbf{mAP@50} & \textbf{mIoU} \\
\midrule
$r = 32$  & 2.54\% & 0.7963 & 0.8614 \\
$r = 64$  & 4.96\% & \textbf{0.8282} & \textbf{0.8631} \\
$r = 128$ & 9.44\% & 0.7804 & 0.8257 \\

\midrule
\textbf{LoRA Injection Location} & \textbf{Params (\%)} & \textbf{mAP@50} & \textbf{mIoU} \\
\midrule
Visual Encoder + Decoder    & 3.71\%  & 0.7881  & 0.8108 \\
Visual Encoder + Decoder + Feature Enhancer & 3.74\% & 0.7966  & 0.8627 \\
LoGSAM (Ours) & 4.96\% & \textbf{0.8282} & \textbf{0.8631} \\

\bottomrule
\end{tabular}
\end{table}

\subsubsection{Ablation Study:}
The ablation study reveals that moderate LoRA ranks provide the best trade-off between capacity and generalization. Performance improves from $r=32$ to $r=64$, where the model achieves its highest localization accuracy. Further increasing the rank to $r=128$ degrades performance, suggesting overfitting.  Similarly, adapting both visual and cross-modal components yields stronger results than modifying visual layers alone. These findings indicate that selective, low-rank adaptation is sufficient for effective MRI-specific grounding.
LoRA-based adaptation introduces only a modest mAP reduction while preserving stable IoU and reliable box placement to guide MedSAM accurately without fine-tuning. 

\section{Conclusion}
In this work, we demonstrate that LoGSAM effectively leverages pretrained foundation models to perform brain tumor localization and segmentation from radiologist speech cues with minimal task-specific fine-tuning. By integrating speech recognition, clinical NLP, vision-language grounding, and prompt-based segmentation into a unified pipeline, our approach enables a modular tumor-analysis while keeping ~95\% of GDINO's parameters frozen. The framework reduces dependence on densely annotated datasets and shows how pretrained foundation models can be adapted for brain tumor analysis.

Future work will extend the framework to 3D detection and segmentation, incorporate multiple MRI sequences, evaluate on larger and more diverse multilingual dictation sets, and explore richer prompt representations that better exploit the semantic content of radiologist speech.

%
%

\end{document}